\documentclass[letterpaper]{article} % DO NOT CHANGE THIS
\usepackage[preprint]{aaai2027}  % DO NOT CHANGE THIS
\usepackage[hyphens]{url}  % DO NOT CHANGE THIS
\usepackage{graphicx} % DO NOT CHANGE THIS
\usepackage{natbib}  % DO NOT CHANGE THIS AND DO NOT ADD ANY OPTIONS TO IT
\usepackage{caption} % DO NOT CHANGE THIS AND DO NOT ADD ANY OPTIONS TO IT
\usepackage{algorithm}
\usepackage{algorithmic}

\usepackage{newfloat}
\usepackage{listings}
\DeclareCaptionStyle{ruled}{labelfont=normalfont,labelsep=colon,strut=off} % DO NOT CHANGE THIS
\floatstyle{ruled}
\newfloat{listing}{tb}{lst}{}
\floatname{listing}{Listing}

\usepackage{booktabs}

\usepackage{array}
\usepackage{amsfonts}
\usepackage{subcaption}
\usepackage{booktabs}
\usepackage{makecell}
\usepackage{graphicx}
\usepackage{makecell}
\usepackage[table]{xcolor}
\usepackage{caption}
\usepackage{amsmath} 
\usepackage{nameref}

\title{Learning to Orchestrate Vision Foundation Models for Multi-Task Dense Prediction}
\author{
Donghyun Han\textsuperscript{\rm 1},
Yuseok Bae\textsuperscript{\rm 1},
Jung Uk Kim\textsuperscript{\rm 2}\corresponding,
Hyung-Il Kim\textsuperscript{\rm 3}\corresponding
}
\affiliations{
    \textsuperscript{\rm 1} Electronics and Telecommunications Research Institute (ETRI), Korea\\
    \textsuperscript{\rm 2} Kyung Hee University, Korea \\
    \textsuperscript{\rm 3} Chonnam National University, Korea \\
    
    \{mpolio2, baeys\}@etri.re.kr, ju.kim@khu.ac.kr, hyungil.kim@jnu.ac.kr
}

\begin{document}

\maketitle

\begin{abstract}
\label{sec:abs}
Vision foundation models (VFMs) exhibit complementary strengths shaped by their pretraining objectives. Yet prevailing methods for multi-task dense prediction still train an entire backbone, either by fine-tuning it under multi-task supervision or by distilling multiple VFMs in an additional stage. We ask whether downstream learning can instead compose the frozen representations already available in foundation models. Dense tasks require composite representations that no individual expert provides alone. Realizing them is difficult: simple fusion yields only marginal gains over the best single expert, while learned routing tends to collapse toward candidates that are strong at initialization, starving newly initialized composers of training signal. We present COVE, which constructs pairwise composite candidates through Synergy Composers and routes among raw and composite candidates with a Task-Conditioned Router. To prevent this collapse, COVE combines Gaussian logit perturbation for exploration with counterfactual supervision that selectively increases under-credited routing allocations. On NYUD-v2 and PASCAL-Context, COVE matches or surpasses ViT-L-based methods on most tasks using a smaller frozen encoder pool and roughly half the computation of recent VFM-based competitors, while exceeding the best single frozen expert on every task.
\end{abstract}
\section{Introduction}
\label{sec:intro}

\begin{figure}[t]
    \centering
    \includegraphics[width=0.7\columnwidth]{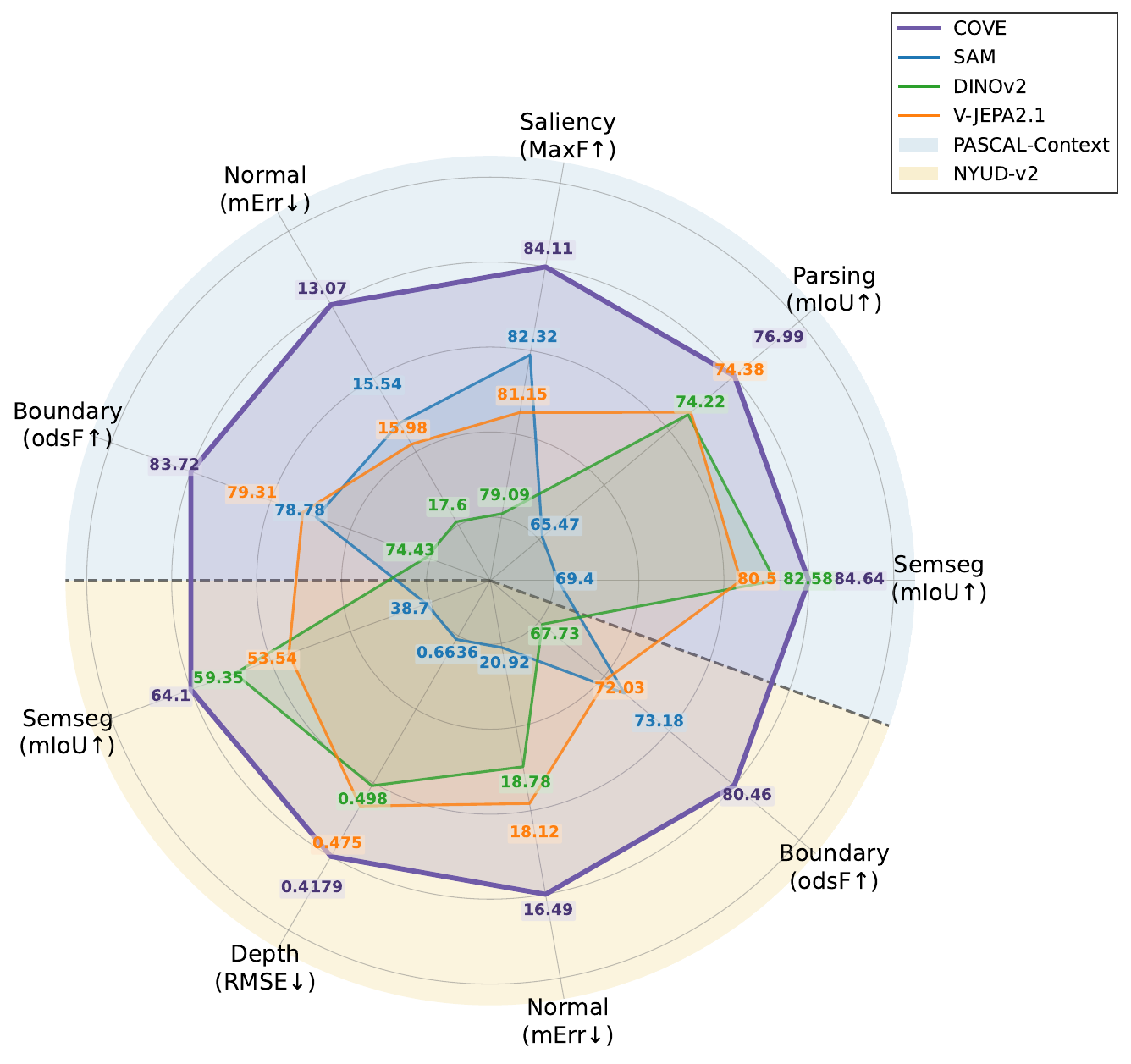}
    \caption{
    Complementary strengths of frozen VFM experts. Each expert excels at different dense tasks, while COVE --- training only lightweight orchestration modules --- surpasses every expert on every task on both NYUD-v2 and PASCAL-Context ($\Delta_\mathrm{E} = +9.74\% $ on NYUD-v2).
    }
    \label{fig:ridar}
\end{figure}

Vision foundation models (VFMs) pretrained at scale now serve as strong feature extractors across a wide range of vision tasks~\cite{sam, dinov2, clip, dinov3}. Their strengths, however, are far from uniform: distinct pretraining objectives induce complementary representations, excelling respectively at semantic understanding~\cite{clip,siglip2}, geometric reasoning~\cite{croco, depthanythingv2}, or fine structural localization~\cite{sam, sam2}. Consequently, no single VFM performs best across the tasks required by multi-task dense prediction, motivating the use of multiple complementary experts over any single model (Figure~\ref{fig:ridar}).

% Existing approaches obtain such broad capability by adapting a unified backbone~\cite{mti-net,invpt,taskprompter,taskexpert}. Conventional multi-task methods fine-tune a large pretrained model under multi-task supervision~\cite{invpt,taskprompter,taskexpert,mlore}, while more recent approaches distill multiple VFMs into a single student, either as a general-purpose backbone~\cite{radio} or for multi-task dense prediction~\citep{sak}, achieving strong downstream performance. Such distillation-based approaches additionally require a large-scale training stage before downstream adaptation. Both paradigms incur substantial training cost and growing model size. Moreover, distillation approximates the experts' knowledge rather than leveraging their representations directly, and the specialized strengths of heterogeneous experts may not be fully preserved when compressed into a shared student~\cite{sak}. The complementary representations already encoded in frozen VFMs therefore remain largely untapped. This naturally raises a more fundamental question: now that diverse visual representations can be acquired through large-scale pretraining, must downstream multi-task learning still adapt a backbone, or can it instead learn only how to compose representations already available?

Existing approaches obtain such broad capability by adapting a unified backbone. Conventional multi-task methods fine-tune a large pretrained model under multi-task supervision~\cite{invpt,taskprompter,taskexpert,mlore}, while more recent approaches distill multiple VFMs into a single student~\cite{radio, sak}. Both paradigms incur substantial training costs and increase model size. Moreover, because a distilled student approximates rather than directly consumes the experts' representations, preserving all expert-specific strengths can be challenging~\cite{sak}. This raises a fundamental question: now that diverse visual representations are already available from large-scale pretraining, must downstream multi-task learning still adapt a backbone, or can it instead learn only how to compose them?

% We study this question under a deliberately constrained setting in which every expert remains entirely frozen and only lightweight composition modules are trained on the target tasks. Under this setting, simply selecting the best expert for each task is insufficient, as dense prediction often requires composite representations unavailable from any single expert. We therefore refer to what must be learned as \emph{orchestration}: constructing composite representations from frozen experts, and adaptively selecting and aggregating raw and composite candidates per task and per spatial location, while every expert remains untouched. Orchestration thus subsumes expert selection but goes beyond it, as the pool of representations to select from is itself expanded.

We study this question under a deliberately constrained setting in which every expert remains frozen. Simply selecting the best expert for each task is insufficient, as dense prediction often requires composite representations unavailable from any single expert. We refer to this as \emph{orchestration}: lightweight modules construct composite representations from frozen experts and adaptively aggregate raw
and composite candidates for each task and spatial location. This goes beyond expert selection by expanding the representation pool itself.

% Realizing such orchestration is difficult for two reasons. First, static fusion cannot determine which complementary cues should be combined for each task and location. Simple averaging or concatenation yields only marginal gains over the best single expert(see Section~\nameref{sec:ablation}), showing that expert complementarity is not automatically converted into useful composite representations. Second, learned composition is vulnerable to routing collapse~\cite{thor, xmoe}. Because frozen experts cannot adapt to compensate for poor early routing decisions, the lightweight composition mechanism must learn both expert selection and credit assignment on its own. As a result, initially strong candidates attract disproportionate routing weights and gradients, starving newly initialized composers and creating a self-reinforcing loop in which feature dominance is mistaken for true contribution.

Realizing such orchestration is difficult for two reasons. First, static fusion cannot determine which cues to combine for each task and location: simple averaging or concatenation yields only marginal gains over the best single expert (see the \nameref{sec:ablation} section), indicating that expert complementarity does not automatically translate into useful composite representations. Second, learned composition is vulnerable to routing collapse~\cite{thor, xmoe}. Because frozen experts cannot adapt to compensate for poor early routing decisions, initially strong candidates attract disproportionate routing weights and gradients, starving newly initialized composers and creating a self-reinforcing loop in which feature dominance is mistaken for actual contribution.

% In this paper, we propose COVE (Composition and Orchestration of Vision Experts) to address these difficulties. Our framework comprises Synergy Composers that explicitly construct composite expert representations by pairwise fusion, and a Task-Conditioned Router that dynamically selects among raw and composite expert representations. To prevent routing collapse during training, we further introduce Counterfactual Credit Assignment, which combines Gaussian logit perturbation for exploration with a counterfactual supervision that supervises routing by measured contribution.

In this paper, we propose COVE (Composition and Orchestration of Vision Experts) to address these difficulties. COVE instantiates orchestration with two lightweight modules: Synergy Composers that explicitly construct composite expert representations by pairwise fusion, and a Task-Conditioned Router that dynamically selects and aggregates raw and composite candidates. To prevent routing collapse during training, we further introduce Counterfactual Credit Assignment, which combines Gaussian logit perturbation for exploration with counterfactual supervision that guides routing according to measured contribution.

% Pairwise Synergy Composers construct explicit composite representations through multi-level expert-axis cross-attention, expanding three raw experts into six raw and composite routing candidates. A task-conditioned router selects among these candidates at each token. To prevent routing collapse, Gaussian logit perturbation encourages exploration, while a counterfactual supervision estimates each candidate's contribution through masking-based loss gaps and corrects only under-credited routing allocations. Exploration provides opportunity; counterfactual evidence converts it into credit.

% On NYUD-v2 and PASCAL-Context, COVE with three frozen ViT-B experts matches or exceeds existing ViT-L-based methods on the majority of tasks while using a smaller frozen encoder budget (260M versus 304M for a single ViT-L) and roughly half the computation. COVE also exceeds the expert oracle on every evaluated task ($\Delta_E=+9.74\%$ on NYUD-v2). All results are obtained without expert fine-tuning, distillation, or additional pretraining.

This composition-only regime sacrifices neither capability nor efficiency. On NYUD-v2~\cite{nyud} and PASCAL-Context~\cite{pascal-context}, COVE matches or exceeds ViT-L-based methods on most tasks while using a 260M-parameter frozen encoder pool and roughly half the computation of recent VFM-based competitors. COVE also exceeds the expert oracle---the best single frozen expert selected per task---on every evaluated task ($\Delta_\mathrm{E}=+9.74\%$ on NYUD-v2). All results are obtained without expert fine-tuning, distillation, or additional pretraining.

Our contributions are summarized as follows:

% \begin{itemize}

% \item We formulate multi-task dense prediction as the composition of frozen foundation-model representations, requiring only lightweight composition modules to be trained on downstream tasks.

% \item We propose the Synergy Composer, a lightweight module that constructs composite expert representations unavailable from any single expert by pairwise fusion, expanding the candidate space through multi-level expert-axis cross-attention.

% \item We introduce Counterfactual Credit Assignment for routing, which combines Gaussian logit perturbation for exploration with a counterfactual supervision that selectively corrects under-credited routing allocations, mitigating routing collapse without load balancing and vanishing automatically once routing aligns with measured contribution.

% \item Empirically, COVE demonstrates that composing frozen foundation-model experts is a viable alternative to backbone adaptation, exceeding the expert oracle on every evaluated task while matching state-of-the-art performance at roughly half the computation.

% \end{itemize}

\begin{itemize}

\item We formulate multi-task dense prediction as the orchestration of frozen foundation-model representations, requiring only lightweight modules to be trained on downstream tasks.

\item We propose the Synergy Composer, which constructs composite representations unavailable from any single expert via multi-level expert-axis cross-attention over expert pairs.

\item We introduce Counterfactual Credit Assignment, which combines Gaussian exploration with contribution-based supervisory targets that selectively correct under-allocated routing decisions.

\item COVE exceeds the expert oracle on every evaluated task while matching state-of-the-art performance at roughly half the computation of recent VFM-based competitors, without expert fine-tuning or distillation.

\end{itemize}
\section{Related Work}
\label{sec:related_work}

\subsection{Multi-Task Dense Prediction}
Multi-task dense prediction has evolved from CNN-based architectures, such as PAD-Net~\citep{pad-net} and MTI-Net~\citep{mti-net}, to transformer-based frameworks including ATRC~\citep{atrc}, InvPT~\citep{invpt,invptpp}, TaskPrompter~\citep{taskprompter}, TaskExpert~\citep{taskexpert}, BFCI~\citep{bfci}, 3D-aware learning~\citep{3d-aware}, TSP~\citep{tsp}, and MLoRE~\citep{mlore}. These methods progressively strengthen cross-task interaction through task-aware attention, prompting, or expert specialization. Accordingly, task conditioning is now well established, whether implemented through task prompts~\citep{taskprompter} or task-gated expert mixtures~\citep{taskexpert,mlore}.

% Despite their architectural differences, these methods share a common paradigm: a single backbone is adapted under multi-task supervision, coupling representation learning to the downstream tasks. COVE instead formulates multi-task dense prediction as \emph{representation composition}. Rather than learning new task-specific representations within a shared backbone, it leaves the foundation-model representations themselves untouched and restricts downstream optimization to lightweight composition over a fixed pool of pretrained features. Moreover, unlike prior multi-scale feature fusion that mixes representations along the spatial axis, our composition operates along the expert axis, combining complementary experts at each spatial location.

Despite their architectural differences, these methods share a common paradigm: a single backbone is adapted under multi-task supervision, coupling representation learning to the downstream tasks. COVE instead formulates multi-task dense prediction as the orchestration of frozen representations. Rather than learning new task-specific representations within a shared backbone, it leaves the foundation-model representations themselves untouched and restricts downstream optimization to lightweight orchestration over a fixed pool of pretrained features. Moreover, unlike multi-scale feature interaction across spatial resolutions~\citep{mti-net,invpt}, COVE composes representations along the expert axis at each spatial location.

\begin{figure*}[t]
    \centering

    \begin{subfigure}[b]{0.65\linewidth}
        \centering
        \includegraphics[width=\linewidth]{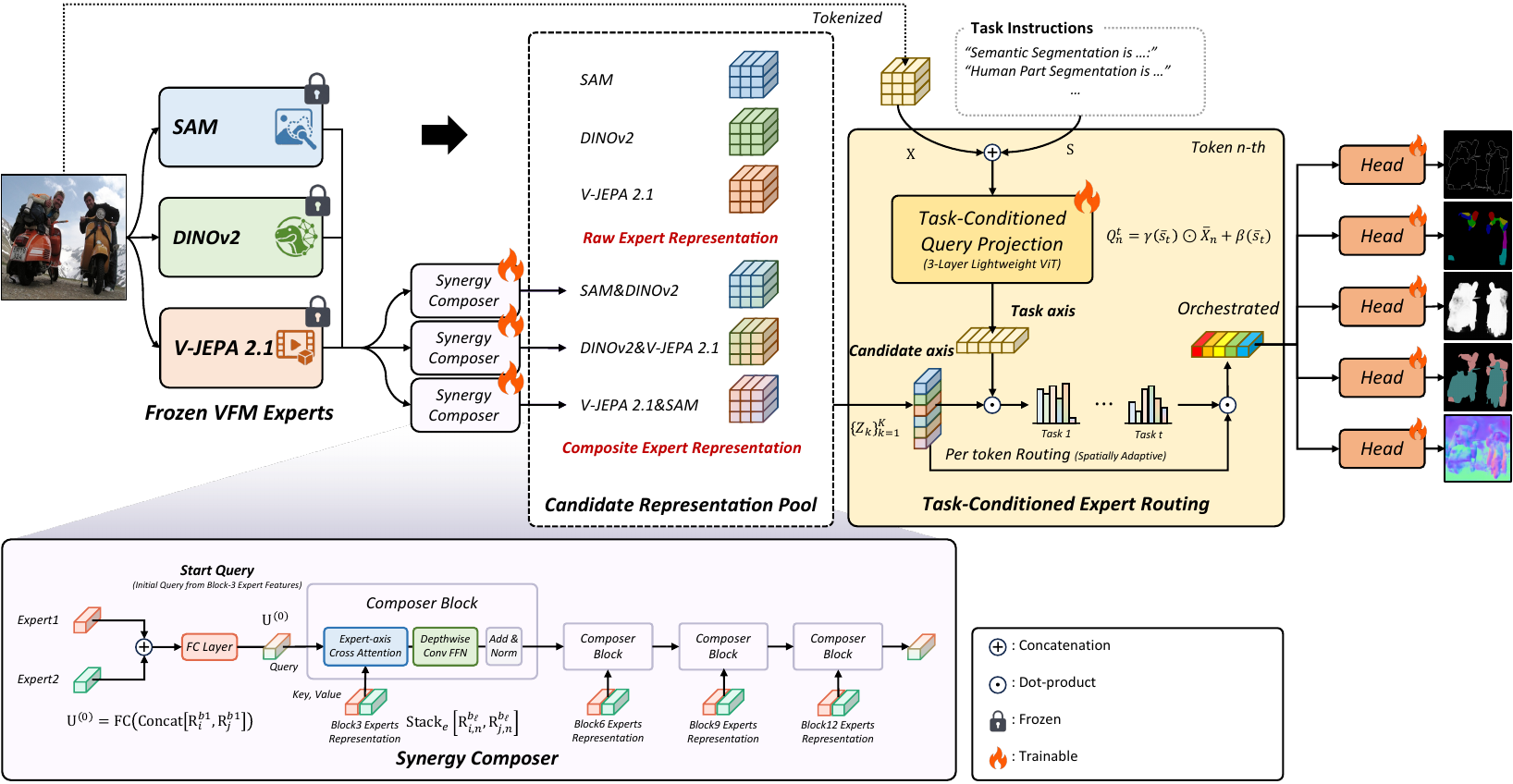}
        \caption{Composition and orchestration of frozen experts}
        \label{fig:overview}
    \end{subfigure}
    \hfill
    \begin{subfigure}[b]{0.34\linewidth}
        \centering
        \includegraphics[width=\linewidth]{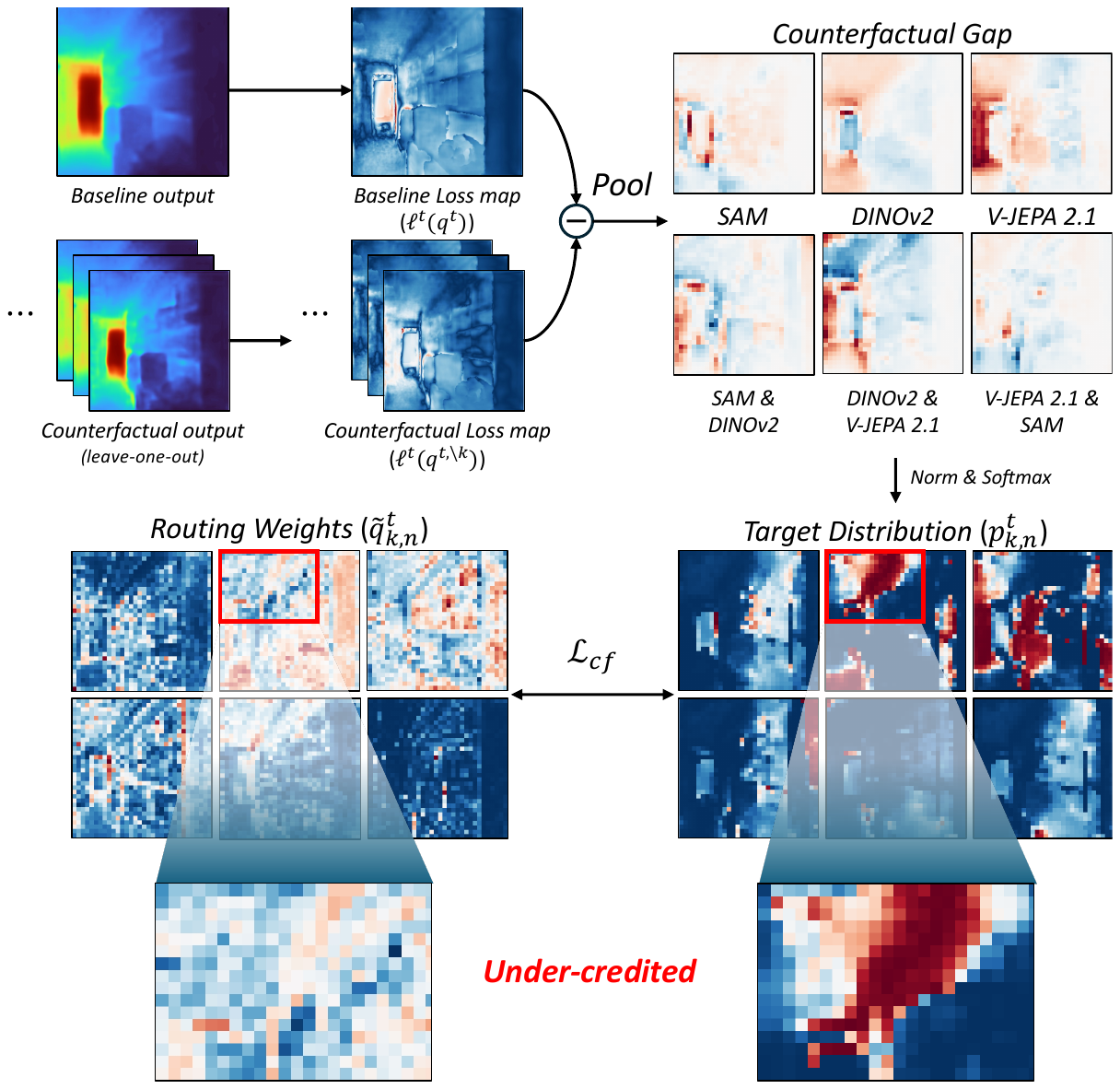}
        \caption{Counterfactual Credit Assignment}
        \label{fig:cf}
    \end{subfigure}

    \caption{Overview of COVE. (a) Synergy Composers pairwise fuse three frozen VFM experts into composite representations, forming a pool of $K{=}6$ candidates from which a Task-Conditioned Router selects per task and per token; only the lightweight orchestration modules are trained. (b) Counterfactual Credit Assignment (training only): leave-one-out loss gaps, measured with the deterministic weights $q$, form a target distribution that pushes up under-credited allocations in the perturbed weights $\tilde{q}$ via $\mathcal{L}_{cf}$.}
    \label{fig:pipeline}

\end{figure*}
\vspace{-2pt}

% \begin{figure*}[t]
%     \centering
%     \includegraphics[width=\textwidth]{Figures/fig_overview.pdf}
%     \caption{}
%     \label{fig:overview}
%     \caption{}
% \end{figure*}

\subsection{Combining Multiple Vision Foundation Models}
Existing approaches to combining multiple VFMs---pretrained with diverse objectives such as promptable segmentation~\citep{sam}, self-distillation~\citep{dinov2}, and predictive modeling~\citep{vjepa_21}---broadly follow two directions.
Distillation-based methods compress multiple experts into a single student. AM-RADIO~\citep{radio} agglomerates heterogeneous VFMs into a general-purpose backbone, while SAK~\citep{sak} extends this paradigm to multi-task dense prediction through a distill-then-adapt pipeline designed to retain teacher-specific biases. Although effective, both require an additional large-scale distillation stage, and the student ultimately approximates the teachers' representations rather than consuming them directly.
A second line of work keeps experts separate and coordinates them through routing. Beyond dense prediction, routing-based coordination has recently been explored for multimodal foundation models, such as MoVA~\citep{mova}. MoVA targets multimodal language understanding, whereas COVE studies task- and token-level routing for multi-task dense prediction.

% COVE follows a different regime. All foundation-model experts remain frozen, so the burden of adaptation falls entirely on the composition mechanism, making routing fundamentally a \emph{credit-assignment} problem rather than a representation-adaptation problem. This departs from conventional mixture-of-experts optimization, where load-balancing objectives regularize aggregate expert utilization~\citep{moe,switch} while individual routing decisions remain unevaluated; concurrent analysis of language-model MoEs reports the same structural gap and corrects it post hoc on a frozen router~\citep{misrouted}. We instead assign credit during training, drawing on counterfactual credit assignment from reinforcement learning, where an agent's contribution is estimated from how the outcome changes when its action is removed~\citep{coma}. Our counterfactual supervision measures token-level predictive contribution through masking-based loss gaps and selectively corrects only under-credited allocations, rather than encouraging balanced utilization.

Unlike these approaches, COVE orchestrates frozen experts: it keeps every expert intact, constructs composite candidates from them, and learns a distribution over raw and composite representations per task and per token. The distinction thus lies in what is learned: distillation re-learns the representations in a student, whereas COVE learns only composition and allocation over untouched expert representations. Consequently, the burden of adaptation falls entirely on the orchestration mechanism, making routing fundamentally a \emph{credit-assignment} problem rather than a representation-adaptation problem. Unlike conventional MoE objectives that regulate token--expert assignment and aggregate utilization~\citep{moe,switch,visionrouters}, COVE directly supervises individual routing decisions by measured contribution. Concurrent work instead corrects misrouted tokens post hoc by updating only the router of a frozen language model~\citep{misrouted}. Inspired by counterfactual credit assignment in reinforcement learning~\citep{coma}, COVE instead assigns routing credit during training through counterfactual supervision, rather than encouraging balanced utilization or correcting routing after the fact.

% This differs from conventional mixture-of-experts optimization, where load-balancing objectives regularize aggregate expert utilization~\citep{moe,switch} while providing no direct supervision for individual routing decisions; concurrent analysis of language-model MoEs reports the same gap and corrects misrouted tokens post hoc by updating only the router of a frozen model~\citep{misrouted}. Inspired by counterfactual credit assignment in reinforcement learning~\citep{coma}, COVE instead assigns routing credit during training through counterfactual supervision, rather than encouraging balanced utilization or correcting routing after the fact.
\section{Proposed Method}
\label{sec:method}
 
% Rather than adapting a backbone to downstream tasks, COVE treats multiple frozen VFMs as complementary experts whose pretrained representations are composed through lightweight learnable modules. Representation extraction and composition are thus explicitly separated: the experts remain fixed, and only the composition modules are optimized on the target tasks.

% Rather than adapting a single backbone to downstream tasks, COVE learns to orchestrate multiple frozen VFMs through lightweight trainable modules. Specifically, orchestration consists of constructing complementary expert representations, dynamically selecting them according to the task, and aggregating them into a unified representation while keeping all experts frozen. 

% Rather than adapting a single backbone to downstream tasks, COVE learns to orchestrate multiple frozen VFMs through lightweight trainable modules. 
% This orchestration is realized through two architectural modules and one training strategy.

Rather than adapting a single backbone to downstream tasks, COVE learns to orchestrate multiple frozen VFMs through lightweight trainable modules, realizing the orchestration defined in the introduction with two architectural modules and one training strategy.

% As shown in Figure~\ref{fig:overview}, we extract representations from three frozen experts---SAM~\citep{sam}, DINOv2~\citep{dinov2}, and V-JEPA\,2.1~\citep{vjepa_21}---whose pretraining objectives induce complementary boundary, semantic, and geometric representations. \emph{Synergy Composers} then fuse each expert pair into a composite representation, forming a pool of $K{=}6$ candidate representations (three raw and three composite). A \emph{Task-Conditioned Router} selects among these candidates per task and per token, and the orchestrated representation is decoded by task-specific heads. Since frozen experts cannot compensate for poor routing through representation adaptation, we train the router with a Counterfactual Credit Assignment strategy (Figure~\ref{fig:cf}).

% As illustrated in Figure~\ref{fig:overview}, COVE first extracts representations from multiple frozen VFMs and augments them with pairwise composite representations generated by \emph{Synergy Composers}, forming a candidate representation pool. A \emph{Task-Conditioned Expert Routing} then dynamically selects and aggregates candidate representations at the token level to produce an orchestrated representation for downstream prediction. During training, COVE further employs a \emph{Counterfactual Credit Assignment} strategy to encourage routing decisions that accurately reflect each candidate representation's contribution. 

As illustrated in Figure~\ref{fig:overview}, COVE first extracts representations from multiple frozen VFMs and augments them with pairwise composite representations generated by \emph{Synergy Composers}, forming a \emph{candidate representation pool}. A \emph{Task-Conditioned Router} then dynamically selects and aggregates candidates from this pool at the token level to produce an \emph{orchestrated representation} for downstream prediction. During training, COVE further employs a \emph{Counterfactual Credit Assignment} strategy to encourage routing decisions that accurately reflect each candidate representation's contribution. 
 
\subsection{Synergy Composer}
\label{sec:composer}
 
\paragraph{Motivation.}
Because the experts are frozen, the router can only select among fixed representations. Yet dense prediction often requires \emph{composite} cues that no single expert encodes---for instance, depth estimation benefits from both geometric structure and precise object boundaries. Selecting the best single expert per task is therefore insufficient (Table~\ref{tab:ablation}), as such composite representations are simply absent from the candidate pool. To supply them, the Synergy Composer explicitly constructs a composite representation from each expert pair, augmenting the candidate pool with both raw and composite candidates.
% To supply them, the Synergy Composer explicitly constructs a composite representation from each expert pair, expanding the candidate pool from three raw experts to six raw and composite candidates.
 
\paragraph{Multi-level expert-axis composition.}
% Given an expert pair $(e_i, e_j)$, the composer consumes their intermediate features at blocks $b \in \mathcal{B} = \{3, 6, 9, 12\}$, denoted $\mathbf{R}_i^{b}, \mathbf{R}_j^{b} \in \mathbb{R}^{N \times d}$, where $N$ is the number of spatial tokens and $d$ the feature dimension; the four blocks uniformly span low- to high-level representations of the shared ViT-B hierarchy at a common token resolution. 

Given an expert pair $(e_i,e_j)$, the Synergy Composer consumes intermediate features from multiple depths rather than only the final block, since cross-expert complementarity may arise at different abstraction levels, from low-level structure to high-level semantics. Let $(b_1,b_2,b_3,b_4)=(3,6,9,12)$ denote the selected expert blocks, with each $b_\ell$ defining composition level $\ell\in\{1,\ldots,4\}$. Their aligned features are denoted by $\mathbf{R}_i^{b_\ell},\mathbf{R}_j^{b_\ell}\in\mathbb{R}^{N\times d}$, where $N$ is the number of spatial tokens and $d$ is the feature dimension.
Since the experts are pretrained with heterogeneous objectives, their features occupy incompatible latent spaces; per-expert alignment MLPs therefore map each expert's features into a shared $d$-dimensional latent space, and all $\mathbf{R}^{b_\ell}$ hereafter denote these aligned features. An initial composition state is formed from the shallowest block,
\begin{equation}
\mathbf{U}^{(0)}
=
\mathrm{FC}\!\left(
\mathrm{Concat}_{c}\!\left[
\mathbf{R}_i^{b_1},
\mathbf{R}_j^{b_1}
\right]
\right),
\end{equation}
where $\mathrm{FC}$ is a fully connected layer and $\mathrm{Concat}_{c}$ denotes channel-wise concatenation. The state is then progressively refined across the four levels in ascending order of depth, integrating low-level cues before higher-level semantics.
 
At each composition level $\ell$, the state attends to the two experts along the \emph{expert axis}: for every spatial token $n\in\{1,\ldots,N\}$, the state token interacts only with its two co-located expert tokens $\{\mathbf{R}_{i,n}^{b_\ell},\mathbf{R}_{j,n}^{b_\ell}\}$ as keys and values,
\begin{equation}
\hat{\mathbf{U}}^{(\ell)}_n
=
\mathbf{U}^{(\ell-1)}_n
+
\mathrm{Attn}\!\left(
\mathbf{U}^{(\ell-1)}_n,\;
\mathrm{Stack}_{e}\!\left[
\mathbf{R}_{i,n}^{b_\ell},
\mathbf{R}_{j,n}^{b_\ell}
\right]
\right),
\end{equation}
% where $\mathbf{U}^{(b-1)}_n$ denotes the state produced at the preceding level in $\mathcal{B}$. This design reflects the purpose of composition---deciding \emph{how to blend} the two experts at each location rather than re-modeling spatial context already captured by the frozen experts---and reduces the attention cost from $\mathcal{O}(N^2)$ to $\mathcal{O}(N)$.
where $\mathrm{Stack}_{e}$ forms two key--value tokens along the expert axis, $\mathrm{Attn}(\mathbf{q},\mathbf{kv})$ denotes cross-attention from query $\mathbf{q}$ to these tokens, and $\mathbf{U}^{(\ell-1)}_n$ is the state produced at the preceding composition level. This design reflects the purpose of composition---deciding \emph{how to blend} the two experts at each location rather than re-modeling spatial context already captured by the frozen experts---and reduces the attention cost from $\mathcal{O}(N^2)$ to $\mathcal{O}(N)$. Since expert-axis attention involves no spatial interaction, a depthwise-convolution FFN restores local spatial context at negligible cost:
\begin{equation}
\mathbf{U}^{(\ell)}=\hat{\mathbf{U}}^{(\ell)}+
\mathrm{DWConvFFN}\!\left(\hat{\mathbf{U}}^{(\ell)}\right).
\end{equation}
The final output $\mathbf{C}_{ij}=\mathbf{U}^{(4)}\in\mathbb{R}^{N\times d}$ is the composite representation for the pair. We instantiate one composer per pair with independent weights, since each pair composes qualitatively different cues.
 
\paragraph{Pairwise vs. joint composition.}
We use independent pairwise composers rather than a single joint composer, preserving each expert combination as a distinct candidate that can be selected and credited independently; a comparison is provided in the supplementary.

\subsection{Task-Conditioned Expert Routing}
\label{sec:routing}
 
\paragraph{Task-conditioned queries.}
% The composer enlarges the candidate pool, but it does not decide which candidate should be used at each location.
The Synergy Composer enlarges the candidate pool, but it does not decide which candidate should be used at each location. Different tasks require different combinations of the available candidates, and within a single task, the requirements vary spatially: contours call for boundary-sensitive cues, whereas interiors are better served by semantic or geometric cues. The router must therefore resolve two questions at once---\emph{which task is being solved} and \emph{what is needed at this location}---and produce a distribution over the candidate pool for every task--token pair.
 
We represent task identity using natural-language instructions encoded by a frozen sentence encoder, providing a shared semantic parameterization without introducing a separate router for each task. Let $\mathbf{X} \in \mathbb{R}^{N \times d}$ denote the tokenized input image, $\mathcal{T}$ the set of target tasks, and $\mathbf{S} = [\mathbf{s}_1, \dots, \mathbf{s}_{|\mathcal{T}|}]$ their task instructions encoded by a frozen sentence encoder~\citep{sbert}. The two are concatenated along the token axis and processed jointly by a lightweight three-layer ViT,
\begin{equation}
\left[\,\bar{\mathbf{X}}\,;\,\bar{\mathbf{S}}\,\right] = \mathrm{ViT}\!\left(\left[\,\mathbf{X}\,;\,\mathbf{S}\,\right]\right),
\end{equation}
so that image content and task semantics interact before routing. This joint pass is shared by all tasks and therefore cannot by itself yield task-specific queries. Instead of instantiating a separate router per task, we modulate the contextualized image tokens with the corresponding task token through FiLM~\citep{film},
\begin{equation}
\mathbf{Q}^{t}_{n} = \gamma(\bar{\mathbf{s}}_t) \odot \bar{\mathbf{X}}_{n} + \beta(\bar{\mathbf{s}}_t),
\end{equation}
% where $\gamma(\cdot)$ and $\beta(\cdot)$ are linear projections. 
where $\bar{\mathbf{s}}_t$ is the contextualized token of task $t$ in $\bar{\mathbf{S}}$, $\odot$ denotes element-wise multiplication, and $\gamma(\cdot)$ and $\beta(\cdot)$ are linear projections. The ViT thus captures image--task interactions shared across tasks, while FiLM supplies task-specific conditioning at negligible cost, so all queries $\{\mathbf{Q}^{t}\}_{t \in \mathcal{T}}$ are obtained from a single forward pass through one shared router.
 
\paragraph{Candidate routing.}
% The candidate pool $\{\mathbf{Z}_k\}_{k=1}^{K}$ ($K=6$) collects the three raw and three composite representations at a common token resolution: $\mathbf{Z}_k = \mathbf{R}^{12}_{k}$ for the three raw experts $k \in \{1,2,3\}$, and $\mathbf{Z}_{3+m} = \mathbf{C}_{ij}$ for the $m$-th expert pair. 

The candidate pool $\{\mathbf{Z}_k\}_{k=1}^{K}$ collects all raw and composite representations at a common token resolution ($K{=}6$ in our setting): $\mathbf{Z}_k = \mathbf{R}^{12}_{k}$ for the raw experts, and the remaining entries are the composite representations $\mathbf{C}_{ij}$ of each expert pair. The router scores each candidate by a dot product with the task-conditioned query and normalizes over the candidate axis,
\begin{equation}
z^{t}_{k,n} = \frac{\langle \mathbf{Q}^{t}_{n},\, \mathbf{Z}_{k,n} \rangle}{\tau}, \qquad
q^{t}_{k,n} = \frac{\exp\!\left(z^{t}_{k,n}\right)}{\sum_{k'=1}^{K} \exp\!\left(z^{t}_{k',n}\right)} ,
\end{equation}
where $\tau=\exp(\tau_0)$ is a learnable positive temperature controlling routing sharpness.
The orchestrated representation for task $t$ is the corresponding convex combination of candidates,
\begin{equation}
\mathbf{H}^{t}_{n} = \sum_{k=1}^{K} q^{t}_{k,n}\, \mathbf{Z}_{k,n},
\label{eq:orchestrate}
\end{equation}
which is decoded by a task-specific head. During training, $\mathbf{H}^{t}_{n}$ is formed with the perturbed weights $\tilde{q}^{t}_{k,n}$ defined in the \nameref{sec:cf} section; the deterministic $q^{t}_{k,n}$ of Eq.~\eqref{eq:orchestrate} is used at inference and, separately, for the counterfactual measurement. Note that this operation never modifies the expert representations themselves: adaptation to downstream tasks is expressed entirely as a choice among pretrained representations. Since each query attends to only $K$ candidates at its own location, routing costs $\mathcal{O}(|\mathcal{T}|NK)$ rather than $\mathcal{O}(N^2)$ and adds little over the frozen experts.

% Since each query attends to only $K{=}6$ candidates at its own location, routing costs $\mathcal{O}(|\mathcal{T}|NK)$ rather than $\mathcal{O}(N^2)$ and adds little over the frozen experts.
 
% \begin{figure}[t]
% \centering
% \includegraphics[width=0.9\columnwidth]{Figures/fig_cf.pdf}
% \caption{Counterfactual Credit Assignment. Candidate contributions are estimated by leave-one-out evaluation and used to supervise routing toward under-credited candidates. Gaps are measured with the deterministic weights $q^{t}$, whereas the supervision is applied to the perturbed weights $\tilde{q}^{t}$ used in the training forward pass.}
% \label{fig:cf}
% \end{figure}

\subsection{Counterfactual Credit Assignment}
\label{sec:cf}
 
\paragraph{Routing collapse.}
In conventional mixture-of-experts training, routing errors are largely self-correcting: an expert that receives little weight can still adapt its representation, and the router's early mistakes are absorbed by representation learning. Freezing the experts removes this mechanism, so credit assignment becomes the learning problem rather than a side effect of it. The difficulty is visible at initialization: raw expert representations are immediately informative, whereas newly initialized composite candidates are not, so the task loss initially favors raw candidates. Because composers receive task gradients through their routing weights, an under-allocated composer receives too little signal to become competitive, while the decoder increasingly specializes to the candidates already favored by the router. We call this self-reinforcing failure \emph{routing collapse}: routing converges according to initial feature strength rather than eventual contribution. Our response is to stop treating routing as something the task gradient shapes implicitly. Instead, we give every candidate the opportunity to be used, explicitly measure its contribution to the prediction, and supervise routing with that measurement rather than with routing probability itself.
 
\paragraph{Gaussian logit perturbation.}
Following noisy routing in mixture-of-experts models~\citep{moe}, we perturb the routing logits with isotropic Gaussian noise before normalization,
\begin{equation}
\tilde{z}^{t}_{k,n} = z^{t}_{k,n} + \xi^{t}_{k,n}, \qquad \xi^{t}_{k,n} \sim \mathcal{N}(0, \sigma^2),
\end{equation}
and use $\tilde{q}^{t}_{k,n} = \mathrm{softmax}_k(\tilde{z}^{t}_{k,n})$ in place of $q^{t}_{k,n}$ throughout training; noise is disabled at inference, where $\tilde{q}$ reduces to $q$. Early in training, perturbation gives every candidate non-negligible routing mass and task gradient, while routing becomes more decisive as the learned logit gaps grow. Perturbation alone, however, only delays collapse: it grants opportunity but carries no information about which candidate deserves weight.
 
\paragraph{Supervising routing by measured contribution.}
The quantity we need is not how strongly the router currently prefers a candidate, but how much the prediction would suffer without it. We obtain it by removing the candidate and observing the effect directly, using the deterministic routing weights $q^{t}$ so that the measurement isolates the effect of removing a candidate from that of the exploration noise. Let $\ell^{t}(\cdot)$ denote the per-pixel task-$t$ loss evaluated on the representation orchestrated by a given weighting; gaps and targets are computed independently per sample, and the batch dimension is omitted for clarity. Masking candidate $k$ , we renormalize the remaining weights using a numerically stabilized denominator:
\begin{equation}
q^{t,\setminus k}_{j,n}
=
\frac{q^t_{j,n}}
{\max\left(\sum_{i\neq k} q^t_{i,n},\,\epsilon\right)}
\quad (j\neq k),
\qquad
q^{t,\setminus k}_{k,n}=0,
\end{equation}
where the denominator is clamped below by $\epsilon$ for numerical stability. The resulting loss difference is pooled to token resolution to obtain the \emph{counterfactual gap}
\begin{equation}
\Delta^{t}_{k,n} = \mathrm{Pool}_{n}\!\left(\ell^{t}\!\left(q^{t,\setminus k}\right) - \ell^{t}\!\left(q^{t}\right)\right),
\end{equation}
computed without gradients, where $\mathrm{Pool}_{n}$ averages the per-pixel difference over the region corresponding to token $n$. A large gap means the prediction at that token genuinely depends on candidate $k$; a gap near zero means the candidate is dispensable there. Since the experts, composers, and router are unaffected by masking, only the task-specific head is recomputed: once for the unmasked reference $\ell^{t}(q^{t})$ and once per masked candidate, giving $K{+}1$ head evaluations in addition to the one used for the task loss.
 
The gaps are converted into a target distribution over candidates,
\begin{equation}
p^{t}_{k,n} = \mathrm{softmax}_k\!\left(\frac{\Delta^{t}_{k,n}}{\kappa \cdot \mathrm{std}_k\!\left(\Delta^{t}_{\cdot,n}\right) + \epsilon}\right),
\end{equation}
where the standard deviation over the candidate axis makes the target invariant to the scale of the task loss, $\kappa$ controls its sharpness, and $\epsilon$ is a small constant for numerical stability. This target serves as explicit supervision for the perturbed routing distribution actually used in the training forward pass. Rather than matching it everywhere, we penalize only the candidates whose measured contribution exceeds the weight they were given,
\begin{equation}
\mathcal{L}_{\mathrm{cf}} = \frac{1}{|\mathcal{T}|N} \sum_{t \in \mathcal{T}} \sum_{n=1}^{N} \sum_{k=1}^{K} \left[\, p^{t}_{k,n} - \tilde{q}^{t}_{k,n} \,\right]_{+} \cdot \left(-\log \tilde{q}^{t}_{k,n}\right),
\label{eq:cf}
\end{equation}
where $[\,x\,]_{+} = \max(0, x)$. Both the target $p$ and the deficit coefficient $[\,p^{t}_{k,n} - \tilde{q}^{t}_{k,n}\,]_{+}$ are detached, so gradients reach the router only through $-\log \tilde{q}^{t}_{k,n}$: each under-credited candidate is pushed up with a strength proportional to its measured deficit. The asymmetric deficit has three useful properties: only allocations below their counterfactual targets are corrected; candidates with smaller measured contributions receive less target mass than more consequential candidates; and the penalty vanishes as $\tilde{q}$ approaches $p$, requiring no annealing schedule.
 
\paragraph{Training objective.}
The full objective adds $\mathcal{L}_{\mathrm{cf}}$ to the multi-task loss,
\begin{equation}
\mathcal{L} = \mathcal{L}_{\mathrm{task}} + \lambda_{\mathrm{cf}} \, \mathcal{L}_{\mathrm{cf}},
\end{equation}
where $\mathcal{L}_{\mathrm{cf}}$ is enabled after a short warmup---during which perturbation alone lets every candidate begin to train---and is applied every other step thereafter to amortize the $K{+}1$ additional head evaluations; hyperparameters are given in the implementation details. Like load-balancing objectives, $\mathcal{L}_{\mathrm{cf}}$ is a cross-entropy on the routing distribution; what differs is not the form of the objective but the supervisory signal from which it is constructed---utilization versus measured contribution. This matters precisely because the experts are frozen: balancing utilization does not make an under-credited candidate useful, whereas evidence of its contribution does. The two mechanisms are also complementary, and neither suffices alone: without perturbation, a candidate that is never selected stays untrained, so masking it changes nothing and yields no evidence to act on; without $\mathcal{L}_{\mathrm{cf}}$, exploration is eventually overwhelmed by the accumulated logits. Exploration provides the opportunity, and counterfactual evidence converts it into credit.
% Table: 
% 1. PASCAL
% 2. NYUDv2
% 3. Ablation
% 4. CF vs. load balancing vs. entropy
% 5. Strong Expert (DINOv3, Depth anything v2) 교체
% 6. Composer vs 유사 파라미터 다른 연산과 비교 (MLP, Conv 등 단순 구조) - 후보

% Figure
% 1. teaser
% 2. overview
% 3. cf
% 4. Pascal
% 5. NYUD
% 6. (a) routing 비교 / (b) candidate, routing 후 Feature cluster 비교

\begin{table*}[t]
\centering
\caption{Comparison on PASCAL-Context across trainable multi-task methods, frozen VFM baselines, and COVE.}
\label{tab:pascal}

\setlength{\tabcolsep}{1.5pt}
\renewcommand{\arraystretch}{0.75}

\resizebox{\textwidth}{!}{%
\begin{tabular}{l|cc|cc|ccccc}
\toprule
Method & Backbone & \makecell{Enc.\\(M)} & \makecell{Params\\(M)} & \makecell{MACs\\(G)} & \makecell{Semseg\\mIoU$\uparrow$} & \makecell{Parsing\\mIoU$\uparrow$} & \makecell{Saliency\\maxF$\uparrow$} & \makecell{Normal\\mErr$\downarrow$} & \makecell{Boundary\\odsF$\uparrow$}\\
\midrule

\rowcolor{gray!15}\multicolumn{10}{c}{\textit{Conventional Multi-task Methods}}\\

PAD-Net{\scriptsize~\cite{pad-net}} & ViT-L & 304M & 340M & 871 & 78.01 & 67.12 & 79.21 & 14.37 & 72.60\\
MTI-Net{\scriptsize~\cite{mti-net}} & ViT-L & 304M & 851M & 774 & 78.31 & 67.40 & 84.75 & 14.67 & 73.00\\
ATRC{\scriptsize~\cite{atrc}} & ViT-L & 304M & 330M & 773 & 77.11 & 66.84 & 81.20 & 14.23 & 72.10\\
InvPT{\scriptsize~\cite{invpt}} & ViT-L & 304M & 423M & 672 & 79.03 & 67.61 & 84.81 & 14.15 & 73.00\\
InvPT++{\scriptsize~\cite{invptpp}} & ViT-L & 304M & 421M & 667 & 80.22 & 69.12 & 84.74 & 13.73 & 74.20\\
TaskPrompter{\scriptsize~\cite{taskprompter}} & ViT-L & 395M & 401M & 497 & 80.89 & 68.89 & 84.83 & 13.72 & 73.50\\
TaskExpert{\scriptsize~\cite{taskexpert}} & ViT-L & 304M & 420M & 622 & 80.64 & 69.42 & 84.87 & 13.56 & 73.30\\
BFCI{\scriptsize~\cite{bfci}} & ViT-L & 304M & 477M & 645 & 80.64 & 70.06 & 84.64 & 13.82 & 72.96\\
3D-aware{\scriptsize~\cite{3d-aware}} & ViT-L & 430M & 430M & 670 & 79.53 & 69.12 & \underline{84.94} & 13.53 & 74.00\\
TSP{\scriptsize~\cite{tsp}} & ViT-L & 304M & 423M & 2159 & 81.48 & 70.64 & 84.86 & 13.69 & 74.80\\
MLoRE{\scriptsize~\cite{mlore}} & ViT-L & 304M & 407M & 827 & 81.41 & 70.52 & 84.90 & \underline{13.51} & 75.42\\
TaskDiffusion{\scriptsize~\cite{taskdiffusion}} & ViT-L & 306M & 476M & 1008 & 81.58 & 71.3 & 85.05 & 13.43 & 76.07 \\
AM-RADIO{\scriptsize~\cite{radio}} & ViT-L & 320M & 372M & 1141 & 81.11 & 71.50 & \textbf{85.17} & 13.49 & 74.80\\
SAK{\scriptsize~\cite{sak}} & ViT-L & 354M & 406M & 1193 & \underline{84.01} & \textbf{76.99} & 84.65 & 13.82 & \underline{76.27}\\

\midrule

\rowcolor{gray!15}\multicolumn{10}{c}{\textit{Vision Foundation Model Baselines}}\\

SAM-B/16{\scriptsize~\cite{sam}} & ViT-B & 87M & 181M & 287 & 69.40 & 65.47 & 82.32 & 15.54 & 78.78\\
DINOv2-B/14{\scriptsize~\cite{dinov2}} & ViT-B & 87M & 181M & 282 & 82.58 & 74.22 & 79.09 & 17.60 & 74.43\\
V-JEPA~2.1-B/16{\scriptsize~\cite{vjepa_21}} & ViT-B & 86M & 181M & 282 & 80.50 & 74.38 & 81.15 & 15.98 & 79.31\\

\midrule

\textbf{COVE-B3/16} & ViT-B$\times$3 & \makecell{260M\\[-4pt]{\scriptsize(frozen)}}& 405M & 576 & \textbf{84.64} & \textbf{76.99} & 84.11 & \textbf{13.07} & \textbf{83.72}\\

\bottomrule
\end{tabular}%
}
\end{table*}
\vspace{-2pt}

% ============================================================================

\begin{table}[t]
\centering
\caption{Comparison on NYUD-v2. Backbones are ViT-L for conventional methods, ViT-B for foundation model baselines, and ViT-B$\times$3 for COVE.}
\label{tab:nyud}

\setlength{\tabcolsep}{1.5pt}
\renewcommand{\arraystretch}{0.82}

\resizebox{\columnwidth}{!}{%
\begin{tabular}{l|ccc|cccc}
\toprule
Method & \makecell{Enc.\\(M)} & \makecell{Params\\(M)} & \makecell{MACs\\(G)} & \makecell{Semseg\\mIoU$\uparrow$} & \makecell{Depth\\RMSE$\downarrow$} & \makecell{Normal\\mErr$\downarrow$} & \makecell{Boundary\\odsF$\uparrow$}\\
\midrule

\rowcolor{gray!15}\multicolumn{8}{c}{\textit{Conventional Multi-task Methods}}\\

InvPT{\scriptsize~\cite{invpt}} & 304M & 402M & 601 & 53.56 & 0.5183 & 19.04 & 78.10\\
InvPT++{\scriptsize~\cite{invptpp}} & 304M & 400M+ & 540+ & 53.85 & 0.5096 & 18.67 & 78.10\\
TaskPrompter{\scriptsize~\cite{taskprompter}} & 492M & 513M & 840 & 55.30 & 0.5152 & 18.47 & 78.20\\
TaskExpert{\scriptsize~\cite{taskexpert}} & 304M & 400M+ & 560+ & 55.35 & 0.5157 & 18.54 & 78.40\\
BFCI{\scriptsize~\cite{bfci}} & 304M & 400M+ & 580+ & 55.51 & 0.4930 & 18.47 & 78.22\\
3D-aware{\scriptsize~\cite{3d-aware}} & 304M & 409M & 600+ & 54.87 & 0.5006 & 18.55 & 78.30\\
TSP{\scriptsize~\cite{tsp}} & 304M & 402M & 1697 & 55.39 & 0.4961 & 18.44 & 77.50\\
MLoRE{\scriptsize~\cite{mlore}} & 304M & 551M & 1406 & 55.96 & 0.5076 & 18.33 & 78.43\\
TaskDiffusion{\scriptsize~\cite{taskdiffusion}} & 307M & 750M & 1920 & 56.66 & 0.5033 & 18.13 & 78.89 \\
AM-RADIO{\scriptsize~\cite{radio}} & 320M & 362M & 970 & 59.32 & 0.4698 & 17.46 & 79.41\\
SAK{\scriptsize~\cite{sak}} & 352M & 394M & 1018 & \underline{63.18} & \underline{0.4313} & \textbf{16.25} & \underline{79.43}\\

\midrule

\rowcolor{gray!15}\multicolumn{8}{c}{\textit{Vision Foundation Model Baselines}}\\

SAM-B/16{\scriptsize~\cite{sam}} & 87M & 163M & 254 & 38.70 & 0.6636 & 20.92 & 73.18\\
DINOv2-B/14{\scriptsize~\cite{dinov2}} & 87M & 163M & 249 & 59.35 & 0.4980 & 18.78 & 67.73\\
V-JEPA~2.1-B/16{\scriptsize~\cite{vjepa_21}} & 86M & 163M & 249 & 53.54 & 0.4750 & 18.12 & 72.03\\

\midrule

\textbf{COVE-B3/16} & \makecell{260M\\[-4pt]{\scriptsize(frozen)}} & 386M & 538 & \textbf{64.10} & \textbf{0.4179} & \underline{16.49} & \textbf{80.46}\\

\bottomrule
\end{tabular}
}

\end{table}
\vspace{-2pt}

% ============================================================================

\begin{table}[t]
\centering
\caption{Ablation on NYUD-v2. $\Delta_\mathrm{E}\%$ is computed relative to the best expert per task, analogously to $\Delta_\mathrm{m}\%$.
}
\label{tab:ablation}

\setlength{\tabcolsep}{3pt}
\renewcommand{\arraystretch}{0.9}

\resizebox{\columnwidth}{!}{%
\begin{tabular}{c|c|c|c|cccc|cc}
\toprule
Router & \makecell{Synergy\\Composer} & $\mathcal{L}_{cf}$ & Noise & \makecell{Semseg\\mIoU$\uparrow$} & \makecell{Depth\\RMSE$\downarrow$} & \makecell{Normal\\mErr$\downarrow$} & \makecell{Boundary\\odsF$\uparrow$} & $\Delta_\mathrm{m}\%\uparrow$ & $\Delta_\mathrm{E}\%\uparrow$ \\
\midrule

\multicolumn{4}{c|}{\textit{Best Expert per Task (Oracle)}} & 59.35 & 0.4750 & 18.12 & 73.18 & -- & -- \\

\multicolumn{4}{c|}{\textit{Single-task Baseline}} & 60.98 & 0.4472 & 17.89 & 75.10 & -- & -- \\

\midrule

avg & -- & -- & -- & 61.14 & 0.4452 & 17.54 & 75.26 & 0.72 & 3.83 \\
cat & -- & -- & -- & 62.03 & 0.4415 & 17.04 & 76.42 & 2.38 & 5.49 \\
$\checkmark$ & -- & -- & -- & 62.62 & 0.4371 & 17.09 & 77.10 & 3.02 & 6.13 \\
$\checkmark$ & -- & $\checkmark$ & -- & 62.84 & 0.4353 & 17.12 & 78.04 & 3.48 & 6.60 \\
-- & $\checkmark$ & -- & -- & 63.08 & 0.4223 & 17.25 & 78.80 & 4.38 & 7.47 \\
$\checkmark$ & $\checkmark$ & -- & -- & 63.27 & 0.4198 & 17.01 & 79.13 & 5.04 & 8.12 \\
$\checkmark$ & $\checkmark$ & $\checkmark$ & -- & \underline{63.59} & \textbf{0.4174} & \underline{16.82} & \underline{79.60} & \underline{5.73} & \underline{8.80} \\
$\checkmark$ & $\checkmark$ & -- & $\checkmark$ & 63.24 & 0.4210 & 16.98 & 79.32 & 5.07 & 8.15 \\
$\checkmark$ & $\checkmark$ & $\checkmark$ & $\checkmark$ & \textbf{64.10} & \underline{0.4179} & \textbf{16.49} & \textbf{80.46} & \textbf{6.66} & \textbf{9.74} \\

\bottomrule
\end{tabular}
}
\end{table}
\vspace{-2pt}

% ============================================================================

\begin{table}[t]
\centering
\caption{
Comparison of routing supervision objectives on NYUD-v2.
}
\label{tab:routing_objective}

\setlength{\tabcolsep}{4pt}
\renewcommand{\arraystretch}{0.95}

\resizebox{\columnwidth}{!}{%
\begin{tabular}{l|cccc|cc}
\toprule
Method & \makecell{Semseg\\mIoU$\uparrow$} & \makecell{Depth\\RMSE$\downarrow$} & \makecell{Normal\\mErr$\downarrow$} & \makecell{Boundary\\odsF$\uparrow$} & $\Delta_\mathrm{m}\%\uparrow$ & $\Delta_\mathrm{E}\%\uparrow$ \\
\midrule

Baseline (w/o Routing Supervision) & 63.27 & \underline{0.4198} & 17.01 & 79.13 & 5.04 & 8.12 \\

\midrule

Load-balancing \cite{moe} & \underline{63.50} & 0.4209 & \underline{16.94} & \underline{79.54} & \underline{5.31} & \underline{8.40} \\
Uniform target (CE to uniform) & 62.94 & 0.4214 & 17.04 & 78.24 & 4.48 & 7.55 \\
Entropy regularization & 62.83 & 0.4212 & 16.99 & 77.86 & 4.39 & 7.46 \\

\midrule

Counterfactual Credit Assignment & \textbf{64.10} & \textbf{0.4179} & \textbf{16.49} & \textbf{80.46} & \textbf{6.66} & \textbf{9.74} \\

\bottomrule
\end{tabular}%
}
\end{table}
\vspace{-2pt}

% ============================================================================

\begin{table}[t]
\centering
\caption{Effect of replacing and adding vision foundation models on NYUD-v2.}
\label{tab:addition}

\setlength{\tabcolsep}{3pt}
\renewcommand{\arraystretch}{0.9}

\resizebox{\columnwidth}{!}{%
\begin{tabular}{c|c|c|c|cc|cccc|cc}
\toprule
SAM &
DINOv2 &
V-JEPA~2.1 &
\makecell{Depth\\Anything v2} &
\makecell{Params\\(M)} &
\makecell{MACs\\(G)} &
\makecell{Semseg\\mIoU$\uparrow$} &
\makecell{Depth\\RMSE$\downarrow$} &
\makecell{Normal\\mErr$\downarrow$} &
\makecell{Boundary\\odsF$\uparrow$} &
$\Delta_\mathrm{m}\%\uparrow$ &
$\Delta_\mathrm{E}\%\uparrow$ \\
\midrule

$\checkmark$ & $\checkmark$ & $\checkmark$ & -- &
386 & 538 &
64.10 & 0.4179 & 16.49 & 80.46 &
6.66 & 9.74 \\

\midrule

$\checkmark$ & $\checkmark$ & -- & $\checkmark$ &
386 & 538 &
\underline{64.71} & \underline{0.3755} & \underline{16.17} & \textbf{80.83} &
\underline{9.85} & \underline{12.80} \\

$\checkmark$ & $\checkmark$ & $\checkmark$ & $\checkmark$ &
506 & 693 &
\textbf{65.33} & \textbf{0.3743} & \textbf{15.99} & \underline{80.61} &
\textbf{10.35} & \textbf{13.30} \\

\bottomrule
\end{tabular}%
}

\end{table}
\vspace{-2pt}

% ============================================================================

\begin{figure*}[t]
    \centering
    \includegraphics[width=0.9\textwidth]{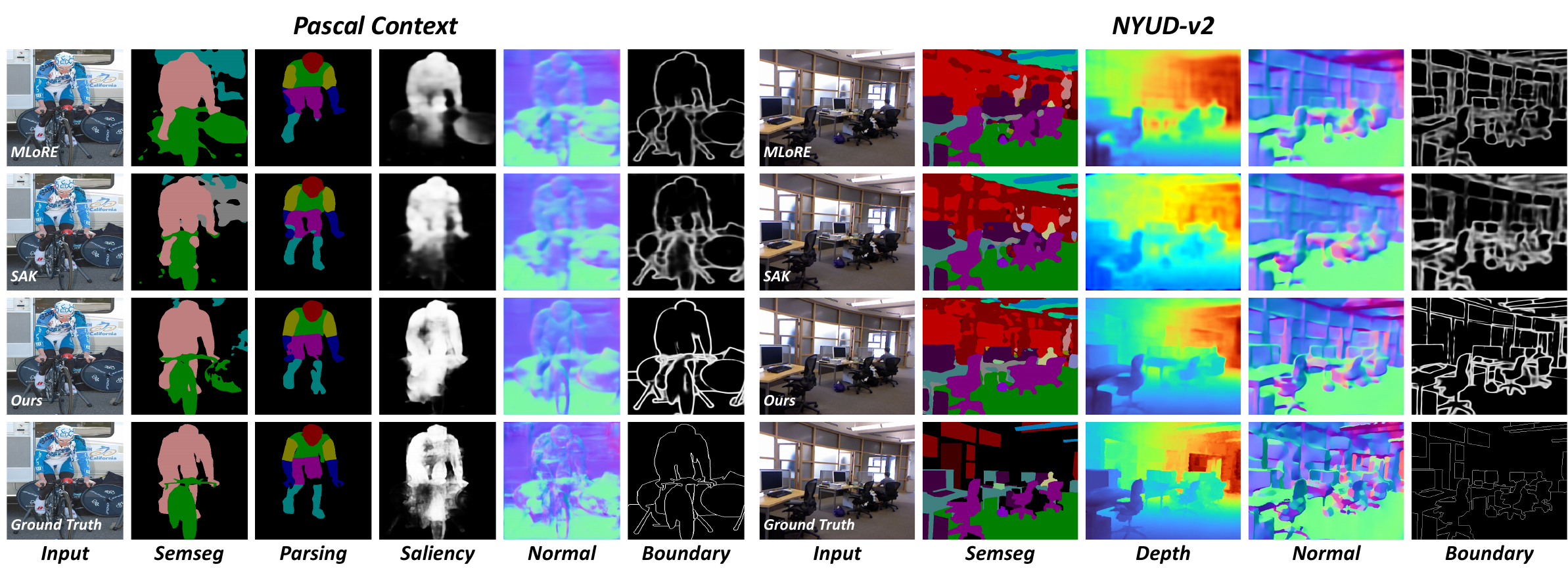}
    \caption{Qualitative comparison with MLoRE~\citep{mlore} and SAK~\citep{sak} on PASCAL-Context (left) and NYUD-v2 (right). COVE produces sharper and more consistent predictions across tasks.}
    \label{fig:comparison}
\end{figure*}
\vspace{-2pt}

% ============================================================================

\section{Experiments}

\subsection{Experimental Setup}

\paragraph{Datasets and metrics.}
We evaluate on NYUD-v2~\cite{nyud} and PASCAL-Context~
\cite{pascal-context} using their standard dense-prediction tasks and metrics. We report $\Delta_\mathrm{m}\%$ relative to single-task models and $\Delta_\mathrm{E}\%$ relative to the best frozen expert per task; the latter measures gains beyond expert selection.

\paragraph{Experts and trainable modules.}
% COVE orchestrates three frozen ViT-B experts—SAM-B/16 \cite{sam}, DINOv2-B/14 \cite{dinov2}, and
% V-JEPA2.1-B/16 \cite{vjepa_21}—totaling 260M frozen encoder parameters.
COVE uses three frozen ViT-B experts---SAM-B/16~\citep{sam},
DINOv2-B/14~\citep{dinov2}, and V-JEPA~2.1-B/16~
\citep{vjepa_21}---totaling 260M encoder parameters. Only alignment layers, composers, the router, and task heads are trained.

\paragraph{Implementation details.}
All models are trained with AdamW following the standard multi-task training protocol used by InvPT~\cite{invpt}, using the same input resolution and schedule across all compared settings. To obtain a common token resolution across experts with different patch sizes, input images are resized per expert (e.g., $512$ for SAM and V-JEPA~2.1 with patch $16$, and $448$ for DINOv2 with patch $14$), yielding an identical $32{\times}32$ token grid. Full hyperparameters, including the perturbation scale $\sigma$, target sharpness $\kappa$, and the loss weight $\lambda_{cf}$, are provided in the supplementary material.

% ============================================================================

\subsection{Comparison with State-of-the-Art}

\paragraph{Results on the PASCAL-Context Dataset.}
We compare our method with state-of-the-art methods built on trainable ViT-L backbones. As shown in Table~\ref{tab:pascal}, COVE achieves the best or tied-best performance on four of the five tasks while keeping all encoders frozen and using roughly half the computation of recent VFM-based competitors. Moreover, COVE surpasses every single frozen expert on every task, indicating that the gains stem from composition rather than any individual representation.

\paragraph{Results on NYUD-v2.}
We also evaluate COVE on NYUD-v2. Without distillation or expert fine-tuning, COVE achieves the best results
on three of four tasks and exceeds the expert oracle by
$\Delta_\mathrm{E}{=}{+}9.74\%$. Qualitative results in Figure~\ref{fig:comparison} further show that our method produces sharper and more consistent predictions.

\subsection{Ablation Study}
\label{sec:ablation}
Table~\ref{tab:ablation} validates each component of COVE on NYUD-v2. Simple fusion of the frozen experts  (avg/cat) yields only marginal gains over the best single expert, confirming that expert complementarity is not automatically converted into useful representations. The Task-Conditioned Router and the Synergy Composer each bring clear improvements, and their combination reaches $\Delta_\mathrm{E}{=}{+}8.12\%$, showing that composite candidates
and adaptive selection are complementary. 
Neither perturbation nor counterfactual supervision is sufficient alone; their combination performs best ($\Delta_\mathrm{E}{=}{+}9.74\%$), confirming their complementary roles.

\paragraph{Comparison of routing supervision.}
Table~\ref{tab:routing_objective} compares $\mathcal{L}_{cf}$ with conventional routing objectives. Load balancing brings marginal gains, while uniform-target and entropy regularization even degrade performance, as they encourage balanced utilization regardless of whether a candidate is useful. In contrast, supervising routing by measured contribution yields the best results on all tasks, validating that credit assignment, not utilization balancing, is the key when experts are frozen.

\paragraph{Effect of expert selection and pool size.}
Table~\ref{tab:addition} shows that COVE benefits from both stronger expert selection and a larger expert pool, reaching \(\Delta_E=+13.30\%\) with four experts.

% ============================================================================

\begin{figure}[t]
    \centering
    \includegraphics[width=0.9\columnwidth]{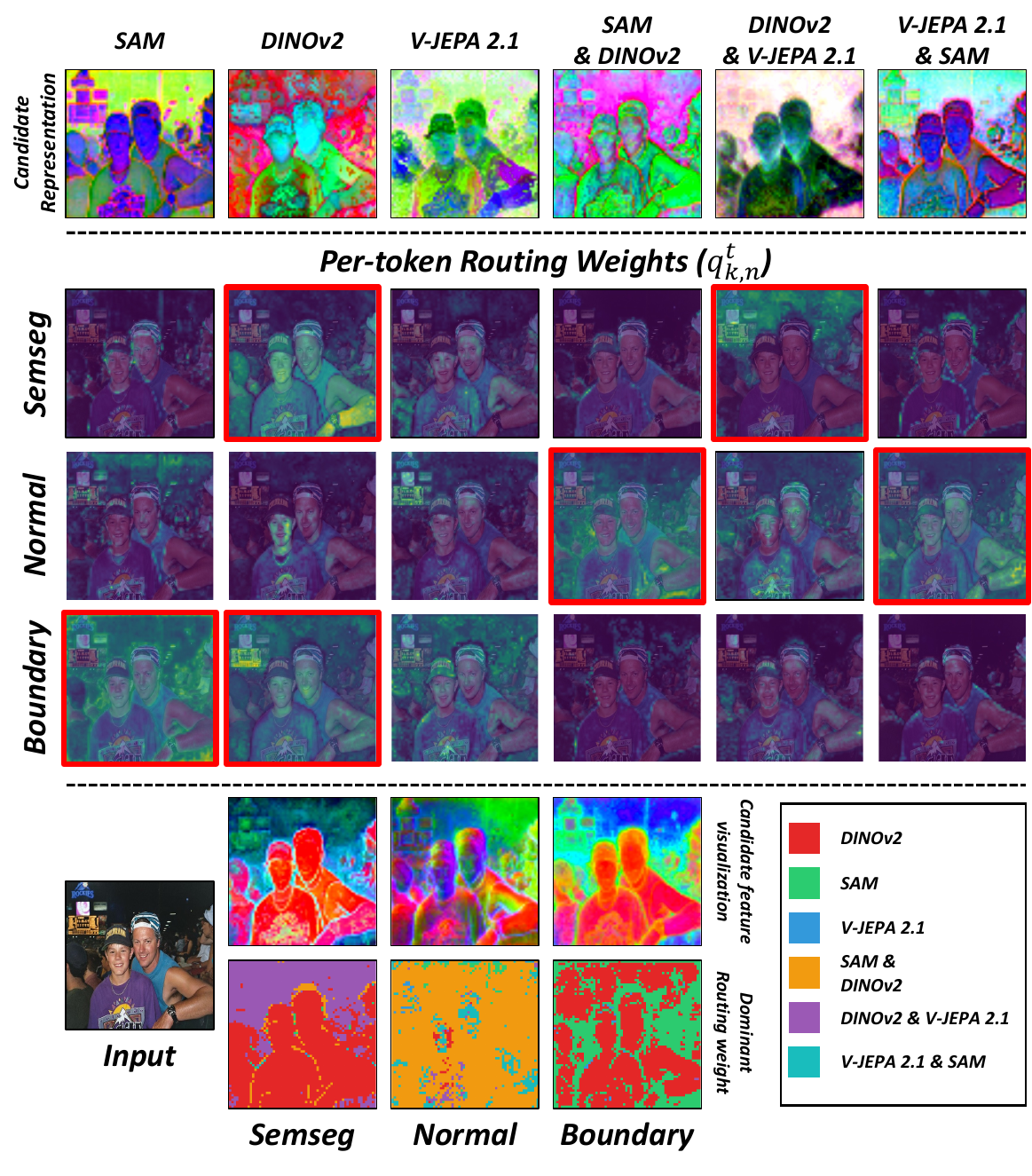}
    \caption{Per-token routing on PASCAL-Context. Routing is task-dependent and spatially adaptive, with substantial weight assigned to composite candidates.
    }
    \label{fig:analysis}
\end{figure}
\vspace{-2pt}

% ============================================================================

\subsection{Discussions}

\paragraph{Routing Analysis.}
Figure~\ref{fig:analysis} visualizes per-token routing on a PASCAL-Context example. Routing is task-conditioned and spatially adaptive: the same image yields distinct dominant candidates per task, shifting between contours and interiors within each task. Notably, composite candidates receive substantial routing mass rather than being starved.

\paragraph{Limitations.}
COVE is bounded by the representations available in its expert pool, although it benefits from stronger experts (Table~\ref{tab:addition}). Counterfactual measurement requires $K{+}1$ additional head evaluations when applied, but is used every other step and adds no inference cost. Finally, task instructions currently provide fixed conditioning rather than adaptation to unseen tasks.

% TODO: 
% teaser(fig1), 
% NYUD, PASCAL 결과 (fig4), 
% Routing, feature cluster analysis (fig5)
% Experiments, Conclusion 채우기
% 지금 section이 ~/ref가 안되는 문제 어떻게 해결할지 생각.

\section{Conclusion}
\label{sec:conclusion}

We asked whether multi-task dense prediction must adapt a backbone, or whether it can instead learn to orchestrate representations that foundation models already provide. COVE realizes this alternative by constructing pairwise composite representations and assigning routing credit according to measured contribution. With all experts frozen, it matches or surpasses trainable ViT-L methods while using roughly half the computation of recent VFM-based competitors, and exceeds the best single expert on every task. As foundation models continue to advance, orchestrating them may become a genuine alternative to adapting them.

\clearpage

\bibliography{aaai2027}

\end{document}